# From Static to Interactive: Adapting Visual in-Context Learners for User-Driven Tasks

**Carlos Schmidt**
Fraunhofer IOSB
Karlsruhe, Germany
carlos.schmidt@iosb.fraunhofer.de

**Simon Reiß** ✉
Karlsruhe Institute of Technology
Karlsruhe, Germany
simon.reiss@kit.edu

## Abstract

Visual in-context learning models are designed to adapt to new tasks by leveraging a set of example input-output pairs, enabling rapid generalization without task-specific fine-tuning. However, these models operate in a fundamentally static paradigm: while they can adapt to new tasks, they lack any mechanism to incorporate user-provided guidance signals such as scribbles, clicks, or bounding boxes to steer or refine the prediction process. This limitation is particularly restrictive in real-world applications, where users want to actively guide model predictions, *e.g.*, by highlighting the target object for segmentation, indicating a region which should be visually altered, or isolating a specific person in a complex scene to run targeted pose estimation. In this work, we propose a simple, yet effective method to transform static visual in-context learners, particularly the DeLVM approach, into highly controllable, user-driven systems, *i.e.*, *Interactive DeLVM* (*i-DeLVM*), enabling seamless interaction through natural visual cues such as scribbles, clicks, or drawing boxes. Specifically, by encoding interactions directly into the example input-output pairs, we keep the philosophy of visual in-context learning intact: enabling users to prompt models with unseen interaction types without fine-tuning and empowering them to dynamically steer model predictions with personalized interactions. Our experiments demonstrate that state-of-the-art visual in-context learning models fail to effectively leverage interaction cues, often ignoring user guidance entirely. In contrast, our method excels in controllable, user-guided scenarios, achieving improvements of $+7.95\%$ IoU for interactive segmentation, $+2.46$ PSNR for directed super-resolution, and $-3.14\%$ LPIPS for interactive object removal. With this, our work bridges the gap between rigid static task adaptation and fluid interactivity for user-centric visual in-context learning, paving the way towards adaptable and dynamically controllable models.

## 1 Introduction

Specialist deep learning models, which are able to solve singular image processing tasks, have been successfully applied in diverse areas. Segmentation models, which are able to delineate specific entities in images are used in medical settings (30), pose estimation models that can extract the orientation of human limbs are used to analyze athletes (5) or image editing models are used in photography to alter image content, *e.g.*, by removing objects (46). Interactive specialist models enable users to steer the output into a specific direction, *e.g.*, by specifying target instances of interest as in medical interactive segmentation (24).

As retraining models for such diverse tasks is costly, due to data annotation and training, techniques which enable adaptation to new tasks with few example images have recently sparked interest. Visual in-Context Learning emerged to address this issue by training models in a few-shot, multi-task setting, for these models to readily learn adaptation to new tasks in training (3; 42). Progress in this direction is accelerating with generative models based on latent diffusion (44; 7) or, more recently, by leveraging sequence prediction models (2; 13). The training paradigm of these approaches is mostly similar: example input-output image pairs $e_i = (c_{in}, c_{out})$ are arranged into a context set $\mathcal{C} = \{e_0, \ldots, e_n\}$ which specify the task to be solved, *e.g.*, for a segmentation task, $c_{in}$ is an image, while $c_{out}$ would be the segmentation mask. Then the visual in-context learning model $\theta(\cdot)$ shall transform a query image $\mathcal{Q}$ analog to the task in the context set: $\theta(\mathcal{C}, \mathcal{Q}) = \mathcal{O}$, *e.g.*, if $\mathcal{C}$ defines segmentation, $\mathcal{O}$ shall be the segmentation of the content in $\mathcal{Q}$. In our research we follow sequence prediction visual in-context learning and analyze whether large vision models can pick up visual interaction signals for better user-centric control.

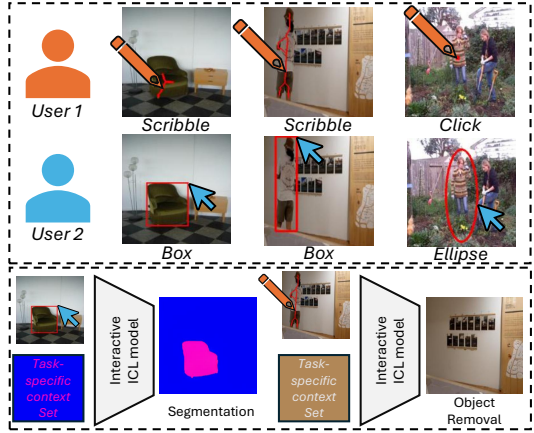

Figure 1: Test-time adaptation to new interactions enables following user preferred interaction signals (top) and allows an adaptive use of interactions to best fit the task at hand (bottom). Enabling a flexible definition of interaction signals at test-time requires vision models that can readily adapt to such different cues.

We find that the existing state-of-the-art models can not successfully generalize to interactive vision tasks (*cf.* Section 5.1), thus, two questions follow: How can we train a visual in-context learning model that can handle *interactive tasks* and that can readily *adapt to unseen interaction cues on the fly*? Addressing these questions can make large vision models more user-centric by allowing the user defined tasks to be interactive in nature and by enabling personalized interaction cues that can be defined by users at test-time.

Figure 1 shows the benefits of models that can handle multiple interaction cues. In the top dashed box, user preferences or user-shift (25) can be accounted for, *e.g.*, if one user prefers free-form drawing cues as interactions and another shape-based selection or in case different users interact slightly differently. The bottom shows that different interaction cues may be better suited for different tasks: for segmenting a compact object such as an armchair a box-interaction might be well suited, while, for editing an image, nuanced scribbles covering each limb of a person to be removed unequivocally defines the removal task.

Importantly, the type of interaction that a specific user might want to use for a given task is not necessarily within the set of interactions used in training the visual in-context model (*e.g.*, training on interactions such as boxes, clicks, scribbles or a circle around target structures), but might extend to other interactions that are not known in advance (*e.g.*, users might want to trace edges of target structures, interact through heatmaps from touch or gaze, draw freeform strokes, *etc.* – or even arbitrary combinations thereof). Therefore, ensuring that models can adapt to unseen interaction cues without retraining is fundamental when bringing interactions to visual in-context learning.

To this end, we propose our blending-based encoding for visual interaction signals which enables sequence prediction-based fine-tuning with interactive tasks. We apply this strategy to the parameter-efficient sequence prediction model DeLVM (13), thereby, turning it into an *Interactive DeLVM* (*i-DeLVM*) variant, and show that it can learn to generalize towards unseen interactions at test-time. Our contributions amount to:

1. We propose a simple and generic visual encoding for interaction cues and a training strategy for turning static visual in-context learning models into interactive models with capabilities to adapt to new interaction cues at test-time.

2. We provide a detailed analysis of visual interactions in the token space of in-context models, including cues such as bounding boxes, circles, clicks and scribbles, measuring their effect on in-context prediction quality.

3. We benchmark state-of-the-art visual in-context learning models and show that they are poorly equipped for interactive tasks, while, our *i-DeLVM* model, which is adapted towards interactivity with our interaction encoding and training strategy, can generalize to unseen interactions, surpassing all prior models on three vision tasks.

## 2 Related Work

### 2.1 Visual in-Context Learning

In natural language processing, it has been found in recent years, that parameter-heavy transformer models (41) pre-trained on large quantities of text exhibit the capability to solve new tasks at inference time by being supplied with few-shot prompts without fine-tuning for the new task (6). This so called in-context learning property has been recognized by the computer vision community as a promising direction to circumventing costly image annotation and model re-training. First efforts of trying to replicate this property were based on inpainting models (3), which formulate the few-shot prompt in an image-grid structure (42).

This grid-based processing was also explored in medical computer vision (32), where the potential to apply in-context learning may help in addressing the notorious data-scarcity issue in the domain. Other variants in the medical domain consider architectures with explicit conditioning on the few-shot examples via a fusion module to enable in-context learning on neuro-images (9) or optical coherence tomography (26).

Another architecture for visual in-context learning uses pre-trained latent diffusion models (33) where the context examples are integrated with ControlNet (47) leading to the Prompt Diffusion approaches (44; 7).

The final modeling choice for visual in-context learning in literature aims to follow natural language processing more closely by employing similar transformer architectures and operating on tokenized images which are put into a sequence structure. Specifically, images are tokenized with Autoencoder models (10) that have a quantization layer (40). Each image then corresponds to a sequence of discrete tokens, which are concatenated to encode the few-shot prompt in one long sequence. Operating on sequences from a diverse set of vision tasks and training a Llama transformer architecture (12) for next-token prediction lead to the Large Vision Models (LVMs) of Bai *et al.* (2). This approach was expanded on with an exploration of data-efficiency and a more parameter-efficient version was obtained through knowledge distillation (13), which is termed DeLVM. Recently, more complex compositional tasks (31) were investigated for visual in-context learning under the utilization of sequence prediction modeling, yet all these previous explorations do not consider interactive tasks. We close this gap and expand on training strategies for DeLVM-style models to advance their capabilities towards interactive vision tasks.

### 2.2 Interactions in Vision and Segmentation

Enabling interactions in computer vision applications has been studied for a long time with interactions such as drawing boxes, clicks or scribbles (34; 27; 4).

With the increased use of deep neural networks, especially interactions in semantic segmentation have been extensively studied, starting with click interactions (45). Subsequent approaches tried to reduce the number of interactions while still producing a high quality image segmentation (39; 21) or refining segmentation iteratively based on falsely predicted regions (15; 23). Following this, interactive segmentation has been improved along many dimensions, such as compute efficiency (37; 38), improving performance via test-time tuning (18) or adapting models to produce a set of diverse segmentation proposals (20).

Recently, interactive segmentation has been scaled to more data and multiple interaction cues (17; 16; 22), broadening the modes of interactions for users. Our work aligns with this, as we also aim to enable multiple interaction cues, yet, we aim to design them in a way to readily adapt to unseen cues. Similar to visual in-context learning, Cheng *et al.* (8) propose interactive segmentation with a context set of images, yet, we explore a broader set of vision tasks and accept a variety of interactions, not exclusively clicks.

## 3 Methodology

Next, we describe the formal framework of interactive visual in-context learning and how we train our interactive in-context model with sequence modeling. Finally, we analyze in detail how a variety of interaction types can be encoded in a unified way and how to utilize them in sequence prediction models. Specifically, we consider the interaction cues boxes, circles, scribbles and two variants of clicks.

### 3.1 Preliminaries

#### 3.1.1 Problem Formulation

Opposed to visual in-context learning, where a context set defines the transfer from an input image $c_{in} \in \mathbb{R}^{3 \times W \times H}$ to a task specific output $c_{out} \in \mathbb{R}^{3 \times W \times H}$, in interactive visual in-context learning, we introduce an additional interaction cue $c_{int}$. As such, each constituent in the context set $\mathcal{C}$ that defines the interactive task to be solved is a triplet $(c_{in}, c_{int}, c_{out})$ instead.

Further, the query $\mathcal{Q}$, which in standard visual in-context learning is just an image, becomes a pair of image and interaction $\mathcal{Q} = (q_{in}, q_{int})$ which needs to be transformed into an output image $\mathcal{O} \in \mathbb{R}^{3 \times W \times H}$. Here, $q_{int}$ stems from a set of interactions $\mathcal{I}$ (*e.g.*, bounding boxes, clicks, *etc.*). Crucially, given the same context set $\mathcal{C}$ and query image $q_{in}$ and a different interaction cue $q_{int}$, the target output $\mathcal{O}$ might change as the interaction might steer the focus for solving a task to a different instance or region in the query image.

The goal of any interactive visual in-context learning model is twofold: (1) the model prediction for the interactive task must correctly predict the target $\mathcal{O}$ and (2) the model must generalize to new interaction types from a set of unseen interactions $\mathcal{I}^{unseen}$, *i.e.*, interactions not present in training. More formally, an interactive visual in-context model $\theta(\cdot)$ shall for a new interaction signal $c_{int'} \in \mathcal{I}^{unseen}$, a context set $\mathcal{C} = \{(c_{in}^0, c_{int'}^0, c_{out}^0), \ldots, (c_{in}^n, c_{int'}^n, c_{out}^n)\}$ and a query $(q_{in}, q_{int'})$, predict the correct task-specific output $\mathcal{O} = \theta(\mathcal{C}, (q_{in}, q_{int'}))$.

#### 3.1.2 Sequence Tokenization

Most visual in-context learning approaches build upon image tokenization based on Vector Quantizated Autoencoders $\phi_{dec}(\cdot) \circ \phi_{enc}(\cdot)$ (40; 10) for more compact image representations. Instead of operating on full resolution images $x \in \mathbb{R}^{3 \times W \times H}$, these tokenizers enable working on quantized encodings $\phi_{enc}(x) \in [0, \ldots, N-1]^{w \times h}$, where $N$ is the number of tokens in a learned codebook and $w \ll W$, $h \ll H$ holds. With the decoder, a token matrix $m \in [0, \ldots, N-1]^{w \times h}$ can be decoded back into image space $\phi_{dec}(m) \in \mathbb{R}^{3 \times W \times H}$.

Existing works either utilize the encoded image in an diffusion process (44) or for auto-regressive next-token prediction (13; 2). We follow the latter approach which requires the formation of a token sequence rather than operating on token matrices of shape $w \times h$. Specifically, we encode all images in the context $\mathcal{C}$ and the query $\mathcal{Q}$ separately with $\phi_{enc}(\cdot)$ into a sequence of tokens $\mathcal{C}_{enc}\mathcal{Q}_{enc} \in [0, \ldots, N-1]^{(|\mathcal{C}|+|\mathcal{Q}|) \cdot h \cdot w}$ through a sequential scanning order for each encoded image. In training, we have access to the single image target $\mathcal{O}$ and thus can encode it into the token sequence $\mathcal{O}_{enc} \in [0, \ldots, N-1]^{h \cdot w}$ in order to train a model to predict it given $\mathcal{C}_{enc}\mathcal{Q}_{enc}$. Similarly, in inference, the visual in-context learner is tasked to predict the tokens $\mathcal{O}_{enc}$ which can be decoded into the correct output image $\phi_{dec}(\mathcal{O}_{enc}) = \mathcal{O}$ based on the input $\mathcal{C}_{enc}\mathcal{Q}_{enc}$.

### 3.2 Interactive Sequential Visual in-Context Learners

To use these sequences for interactive visual in-context learning, we now describe our method for generic interaction encoding and the corresponding training strategy to obtain our interactive visual in-context learning model *i-DeLVM*.

#### 3.2.1 Encoding Visual Interaction Signals

We aim to encode visual interactions such as boxes, clicks, scribbles and circles, all of which might be used to highlight a specific instance in an image. Multiple options for encoding interactions are possible. An interaction-specific encoding considers each interaction separately and encodes it through its specific properties, *e.g.*, boxes and clicks via their coordinates or dense prompts as masks (17), and a successive learned prompt embedding. While proven to work well, this is unsuitable if the goal is to also generalize towards new, unseen visual interactions, as it would be unclear how to encode them without a re-trained prompt embedding.

Thus, we opt for encoding all interaction signals as images $c_{int} \in \mathbb{R}^{3 \times W \times H}$, where pixel-locations with an interaction are marked with an interaction color and all remaining pixels are set to black. With this, we have a homogeneous representation for all visual interactions which allows for inserting unseen interactions $c_{int'}$ at test-time. As our interactive visual in-context inputs $\mathcal{CQ}$ contain $n$ triplets $(c_{in}, c_{int}, c_{out})$ in the context set $\mathcal{C}$ and one in the query $\mathcal{Q} = (q_{in}, q_{int})$, a total of $n+1$ interaction images $c_{int}$ would be added. This in turn, would lead to $(n+1) \cdot w \cdot h$ additional tokens in our token sequences $\mathcal{C}_{enc}\mathcal{Q}_{enc}$, as each interaction image needs to be separately encoded with $\phi(\cdot)_{enc}$.

As this leads to impractically long sequences, especially for the notoriously memory-hungry transformer architecture, we propose to blend $c_{in}$ and $c_{int}$ ($q_{in}$ and $q_{int}$ respectively):

$$c_{in+int} = \mathbb{K}(c_{int} = 0) \cdot c_{in} + (\alpha \cdot \mathbb{K}(c_{int} \neq 0) \cdot c_{in} + (1-\alpha) \cdot \mathbb{K}(c_{int} \neq 0) \cdot c_{int}) , \quad (1)$$

where $\mathbb{K}(\cdot)$ is the indicator function and $\alpha$ is an optional blending factor, which allows for interaction signals to be blended with image content rather than overwrite it, which may be necessary for spacious interaction signals at test-time (*e.g.*, heatmaps). As we consider sparse interaction signals in our setting, we choose $\alpha = 0$ throughout and explore any implications of this choice on overwriting image content in Section 3.3.3. With this collapse of the interaction signal into $c_{in}$, or $q_{in}$ respectively, we effectively keep the sequence length consistent with prior visual in-context models at no added computational cost.

#### 3.2.2 Learning Interactive Tasks with Next Token Prediction

With a formulation of interactive visual in-context learning as token sequences, next, we describe the model training procedure. Therefore, we consider auto-regressive sequence generation as generic way to learn interactive visual tasks with a neural network $\theta$. Specifically, we consider the joint probability of a sequence $\mathcal{O}_{enc} = \{\mathcal{O}^1_{enc}, \ldots, \mathcal{O}^{w \cdot h}_{enc}\}$ as the product of conditional probability distributions:

$$p(\mathcal{O}_{enc}) = \prod_{i=1}^{w \cdot h} p(\mathcal{O}^i_{enc} | \mathcal{C}_{enc}\mathcal{Q}_{enc}, \mathcal{O}^1_{enc}, ..., \mathcal{O}^{i-1}_{enc}; \theta) . \quad (2)$$

Here, contrary to classical auto-regressive training $p(\mathcal{O}_{enc})$ is estimated with the prefix $\mathcal{C}_{enc}\mathcal{Q}_{enc}$, *i.e.*, the context set and query which defines the interactive task. The model $\theta$, which in our case is a transformer (41), is trained to estimate the conditional probabilities for a token $\mathcal{O}^i_{enc}$ through $N$ output values which are fed into the softmax function. The result indicates a distribution for the currently predicted token at position $i$ over all entries $[0, \ldots, N-1]$ in the codebook. During training, the log-probability of the correct token sequences $\mathcal{O}_{enc}$ from the training set with respect to $\theta$ is maximized. We further train by masking different ratios of the tokens $\mathcal{O}^1_{enc}, ..., \mathcal{O}^{i-1}_{enc}$ in Equation (2), to see the effect of the reliance on context tokens $\mathcal{C}_{enc}\mathcal{Q}_{enc}$ versus the previously generated tokens.

In our setup, we utilize pre-trained DeLVM visual in-context learning models (13) and fine-tune for interactive visual tasks by blending interactions with Equation (1) and optimizing conditional probabilities with context and query examples as in Equation (2). Henceforth, we refer to these models as **I**nteractive **D**ata-efficient **L**arge **V**ision **M**odels (*i-DeLVM*s), following Guo *et al.*'s DeLVM naming, although the interaction encoding and training strategy can be generically applied to other visual in-context models. Figure 2 (bottom) shows the tokenization and subsequent next token prediction process to infer an output.

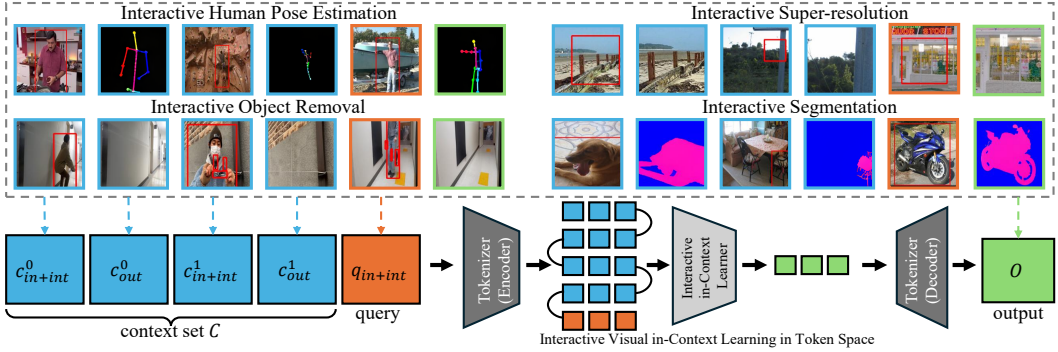

Figure 2: Top: Example context sets, queries and outputs for interactive tasks with blended interactions (*e.g.*, bounding boxes) in input images. Bottom: Processing in *i-DeLVM*. Training with the context set, query and output is conducted in token space.

## 3.3 Visual Interactions and Analysis

While generic, efficient and conceptually simple, blending images and interactions also overwrites visual information from $c_{in}$ (or $q_{in}$), hinting at a trade-off. Next, we introduce different visual interactions and analyze to which extent blending them into images effects reconstruction quality.

### 3.3.1 Visual Interaction Types

We build upon the following five interaction cues:

**Bounding box**: A box drawn tightly around the instance of interest, it encompasses the whole instance as well as some background regions which, for some tasks, are not of interest.

**Ellipse**: Similarly to the box, this interaction has the same properties but delimits the instance in an ellipsoid region instead of a box.

**Scribble**: For this interaction type, a user draws scribble lines onto the instance of interest. All pixels that the scribbles cover belong to the desired instance.

**Click**: This is a special case of a scribble, *i.e.*, it covers just a singular point in the image which lies on the instance.

**Positive negative clicks (± Clicks)**: Similar to the click interaction, here, two types of clicks are possible. The positive click is equivalent to the click interaction above, negative clicks are singular points that explicitly do not lie on the instance of interest, they are encoded as different colors. Further, multiple positive and negative clicks can be present in this interaction cue.

### 3.3.2 Manifestation of Interactions in Token Space

Many visual in-context learning approaches, including our interactive approach operate on tokenized images. In order to make sure, that these often sparse interaction signals do not get lost in the tokenization via $\phi_{enc}(c_{in+int})$ we analyze how well they can be recovered from token space through the de-tokenization process $\phi_{dec}(\cdot)$. As representatives, we look into bounding box and click interactions and blend either onto images from the object-centric PASCAL VOC dataset (11). The interactions are both defined as red pixels, which is in coherence with prior literature that found the color red to work best (36). To measure whether the interaction signals on these images can be recovered, we post-process the auto-encoded blended image $\phi_{dec}(\phi_{enc}(c_{in+int}))$ by extracting the region of the interaction with a 2 pixel tolerance frame around it and detect red color values through a small tolerance distance to the color red. In order to quantify the retention of the interaction signals we measure the mean Intersection over Union (IoU) between the recovered interaction pixels and the original interaction pixels as blended into $c_{in+int}$.

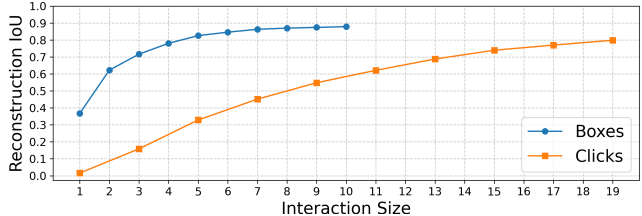

Figure 3: Reconstruction quality of bounding boxes and clicks of different line widths and side lengths, respectively.

| Metric | – | Box | Ellipse | Scribbles | Click | $\pm$ Clicks |
|---|---|---|---|---|---|---|
| SSIM | 62.51 | 61.77 | 60.87 | 61.38 | 62.73 | 62.10 |
| PSNR | 22.23 | 21.33 | 20.94 | 22.40 | 22.18 | 21.62 |

Table 1: Reconstruction quality of images with different interaction types as measured in SSIM ($\uparrow$) and PSNR ($\uparrow$).

In Figure 3, we show how well boxes and clicks can be recovered as function over their thickness. While very thin interactions, *e.g.*, 1 pixel wide interactions, are recovered quite badly, for bounding boxes, a width of 3 pixels leads to a good reconstruction of $> 70\%$ IoU, indicating that the token space conveniently represents boxes with this width. For clicks, due to their extremely sparse nature, the reconstruction improves slower with its size. There, a high reconstruction score and the image content that is occluded by the interaction needs to be balanced. We do not need a particularly precise reconstruction, as we only need to make sure that the interaction signal is present in the token space (for the interactive tasks we do not recover the interaction itself). Thus, we chose a reconstruction of $> 50\%$ IoU, which is reached at an acceptable size of 9 pixels. As such we define boxes, ellipses and scribbles with a width of 3 and clicks as well as $\pm$ clicks as small, 9 pixel wide squares.

### 3.3.3 Interaction-dependent Reconstruction

To better quantify how different interaction signals blended into an image may adversely effect the reconstruction of the image content itself, we measure the image reconstruction quality. Therefore, we autoencode the image blended with the interaction $\phi_{dec}(\phi_{enc}(c_{in+int}))$ and measure its Structural Similarity Index Measure (SSIM) and Peak Signal-to-Noise Ratio (PSNR) compared to the original image $c_{in}$ on the PASCAL VOC dataset.

In Table 1, we see this reconstruction efficacy for different interactions and for the case when the image is not blended with an interaction at all (column two). Generally, boxes, ellipses and scribbles cover more image content and lead to slightly lower image reconstruction quality as opposed to the click variants. As positive and negative clicks cover more image content than singular clicks, they lead to slightly lower scores. All in all, the interactions as we defined them have a rather limited degradation effect on image quality and almost reach up to the quality when no interaction is provided. Interestingly, the SSIM of clicks and the PSNR of scribbles lead to a slightly higher score in reconstruction. This hints at the rather minuscule effect of blending interactions and suggests that better image tokenizer models for a faithful reconstruction of fine-trained details may be an important future research direction.

## 4 Experimental Setup

### 4.1 Datasets and Interactive Tasks

For interactive visual in-context learning, we require multiple datasets which encompass different vision tasks which are augmented with multiple interaction signals. In the following we outline the datasets we train and evaluate on.

**Semantic Segmentation**: PASCAL VOC (11) is a dataset containing 2,913 images with mask annotations of 20 object-classes from daily life and corresponding segmentation masks. The interactive segmentation task resides in predicting the correct segmentation mask of the object or instance which the interaction refers to. We vary the color coding of masks as described in the supplemental material.

**Object Removal**: We use the Real-world Object Removal dataset (35) for interactive object removal, which is the task of erasing the object which the interaction refers to and inpainting the erased region with a texture coherent to the surroundings. The dataset has $3,447$ scenes of which $2,992$ have at least 6 frames per scene, with and without objects, that we use. Images with objects serve as $c_{in}$ and $q_{in}$, images without the objects as $c_{out}$ and $q_{out}$.

**Human Pose Estimation**: The MPII dataset (1) contains $24,984$ images, of which we drop images with extremely small persons that are far away from the camera (*i.e.*, the $15\%$ smallest) to end up at $17,280$ images containing persons with pose annotations, *i.e.*, the orientation of their extremities. We form an interactive task by requiring to exclusively detect the pose of the person which is identified by the interaction cue. The pose output $c_{out}$ and $q_{out}$ is formed as an image with a pose-skeleton using different colors for the bones (*cf.* supplemental material).

**Interactive Super-Resolution**: As a more exotic task, we define interactive super-resolution, which describes the interactive task of selecting a region in an image (with a region based cue, such as boxes and ellipses) which shall be copied and it's image quality shall be improved. Put differently, a region in an image is cropped based on an interaction and subsequently super-resolution is applied. For this task, we select higher quality samples in ADE20K (50) by filtering out images that have widths and heights smaller than 300 pixels, yielding $19,019$ high resolution images.

All tasks are shown in Figure 2, interactions are automatically extracted from accompanying labels in the datasets, all images are resized to $256 \times 256$ for uniform inputs.

### 4.2 Evaluation

We evaluate segmentation with IoU, interactive super-resolution and object removal with SSIM, PSNR and Learned Perceptual Image Patch Similarity (LPIPS) (48). For human pose estimation we show qualitative results, as deconstructing generative image-based pose predictions into metric results such as Percentage of Correct Keypoints (PCK) is still difficult with current state-of-the-art-models. This is due to the sparse joint and skeleton predictions which inherit the limitations of the tokenization process (*i.e.*, tokenizers having problems when reconstructing small, thin structures coherently (31)).

### 4.3 Implementation Details and Baselines

We use the DeLVM VQ-GAN as tokenizer and the DeLVM transformer with 300M parameters, which we fine-tune using LoRA (14) on $0.96\%$ of the parameters (LoRA rank $r = 32$, $\alpha = 32$), tuning the self-attention $Q, V$-projections with dropout of $0.1$. Our initial learning rate is $2.8e^{-4}$ and we use a Cosine Annealing schedule with $T_{\max} = 50$, $\eta_{\min} = 1e^{-5}$. We train with a context window of $2,048$ tokens which amounts to three context pairs $(c_{in+int}, c_{out})$ and one query-output pair $(q_{in+int}, q_{out})$ which are put into a batch of size 8 on an A100 GPU using 16-bit precision. We train by masking $100\%$ of the tokens $\mathcal{O}^1_{enc}, \ldots, \mathcal{O}^{i-1}_{enc}$, when predicting next tokens $\mathcal{O}^i_{enc}$, *i.e.*, relying fully on the context set, except in the ablation study where we vary this masking ratio. As in literature (26), we use random recoloring augmentation, where varying colors are used to encode segmentation tasks, such that models learn to predict in accordance with colors in the context $\mathcal{C}$, rather than overfit to a specific color scheme used in training.

We train our *i-DeLVM* in different settings, the main model, that we are interested in, *i.e.*, a model that adapts to new, unseen interaction cues $\mathcal{I}^{\text{unseen}}$ is what we term *i-DeLVM*$_{\text{Unseen}}$. It is trained on all tasks and all interactions except the interaction that is evaluated, *i.e.*, different models need to be trained with a hold-one-interaction-out scheme for each interaction. This showcases the generalization capabilities to unseen interaction cues.

To compare what the *i-DeLVM* architecture is capable of in general, we add strong specialist variants:

(1) *i-DeLVM*$_{\text{Int}}$ is a model that is trained on all four tasks but one single interaction, as such we train one such interaction-specialist model for each interaction type.

(2) *i-DeLVM*$_{\text{Task}}$ is a model trained on a single task but with all interactions, thus, we train one task-specialist model for each task.

(3) *i-DeLVM*$_{\text{All}}$ is a multi-task multi-interaction model, trained on all tasks and all interaction signals simultaneously.

These three specialists, give insight into supervised model's capabilities and the role of task-/interaction-diversity in training as compared to *i-DeLVM*$_{\text{Unseen}}$. We further evaluate two more baselines, a naive lower baseline and an upper bound based on the image tokenizers, which the visual in-context learning models utilize.

*Copy baseline*: This serves as lower baseline by naively copying either the query image $q_{in}$ or a random target $c^i_{out}$ from the context.

*VQ-GAN Oracle*: This baseline uses the VQ-GAN tokenizer model to auto encode the ground-truth. It shows the best score to be achieved for any in-context learner utilizing this tokenizer.

Finally, we compare to *DeLVM (300M)*, *DeLVM (1B)* and *LVM (3B)* the current state-of-the-art in visual in-context learning. For interactive segmentation, we further evaluate another specialist model, which is *SAM 2* (29).

## 5 Results and Analysis

### 5.1 Quantitative Results

#### 5.1.1 Interactive Semantic Segmentation

Table 2 shows the results for interactive segmentation. The copy target baseline gives a lower bound for the different approaches, while the last row shows the best possible performance of $87.90\%$ IoU when building on top of the VQ-GAN tokenizer. While staying on a low level, pre-trained visual in-context models slightly outperform the copy baseline. It can be seen that the segmentation performance for these models varies only slightly for different interactions, $2.33\%$ for DeLVM (300M), $1.73\%$ for DeLVM (1B) and $0.41\%$ for LVM. The low scores indicate that the interactions are not considered. Our *i-DeLVM*$_{\text{Unseen}}$ model, which did not observe the respective interactions, starts in capturing the interactivity of the task quite well as compared to DeLVM and LVM, with considerable margins of $+12.85\%$, $+10.02\%$, $+9.07\%$, $+7.95\%$ and $+14.64\%$ for the different cues. This shows a clear notion of generalization to unseen interactions. Our variants that were trained on all tasks and all interaction signals, or on individual tasks/interactions – as expected – outperform all other methods, as all of them are informed about the respective interaction cues. The interactive segmentation specialist SAM 2 produces better results for boxes and clicks than these *i-DeLVM* variants, but falls short of *i-DeLVM*$_{\text{Unseen}}$ for ± clicks and does not accept all interactions.

|  |  | Bbox | Ellipse | Scribble | Click | ± Clicks |
|---|---|---|---|---|---|---|
| *Generalist* | Copy target | 5.15 | 5.15 | 5.15 | 5.15 | 5.15 |
|  | DeLVM (300M) | 6.00 | 5.10 | 7.43 | 6.91 | 6.89 |
|  | DeLVM (1B) | 7.94 | 7.81 | 7.72 | 9.31 | 9.45 |
|  | LVM (3B) | 8.73 | 8.32 | 8.54 | 8.65 | 8.63 |
|  | *i-DeLVM*$_{\text{Unseen}}$ | **21.58** | **18.34** | **17.61** | **17.26** | **24.09** |
| *Specialist* | *i-DeLVM*$_{\text{All}}$ | 50.52 | 50.49 | 47.31 | 30.52 | 35.48 |
|  | *i-DeLVM*$_{\text{Int}}$ | 52.91 | **53.67** | **47.40** | 39.19 | **42.16** |
|  | *i-DeLVM*$_{\text{Task}}$ | 35.32 | 33.42 | 36.61 | 24.10 | 26.40 |
|  | SAM 2 | **67.01** | - | - | **45.68** | 15.35 |
|  | VQ-GAN Oracle | 87.90 | 88.05 | 87.35 | 87.58 | 87.77 |

Table 2: Interactive segmentation performance on PASCAL VOC2012 across different interaction cues measured in IoU.

### 5.1.2 Interactive Super-resolution

In Table 3 we show the results for the task of computing a high resolution image for a region selected with a box or ellipse in an input image. As a difficult compositional task, we see that perceptual metrics such as LPIPS start to get smaller only with the supervised *i-DeLVM* models. The baselines and *i-DeLVM*$_{\text{Unseen}}$ struggle in producing perceptually meaningful details. Where we can see a clear separation between baselines and *i-DeLVM*$_{\text{Unseen}}$ is when we look at structural metrics, where the latter captures the most prominent structures, reflected in SSIM scores of $43.01\%$ and $41.67\%$ and generally higher PSNR results for box and ellipse interaction signals, respectively.

|  | | Bounding Box | | | Ellipse | | |
|---|---|---|---|---|---|---|---|
|  | | LPIPS↓ | SSIM↑ | PSNR↑ | LPIPS↓ | SSIM↑ | PSNR↑ |
| *Generalist* | Copy query | **60.25** | 26.01 | 10.43 | 59.01 | 26.24 | 10.52 |
| | DeLVM (300M) | 73.35 | 23.83 | 7.76 | 72.80 | 22.36 | 7.60 |
| | DeLVM (1B) | 61.93 | 24.17 | 9.70 | 61.94 | 23.92 | 9.63 |
| | LVM (3B) | 62.32 | 28.25 | 10.03 | 68.13 | 27.94 | 8.96 |
| | *i-DeLVM*$_{\text{Unseen}}$ | 61.69 | **43.01** | **13.32** | **56.85** | **41.67** | **12.98** |
| *Specialist* | *i-DeLVM*$_{\text{All}}$ | 45.72 | 45.68 | 15.77 | 43.95 | 46.32 | 16.04 |
| | *i-DeLVM*$_{\text{Int}}$ | 45.99 | 47.09 | 16.73 | 43.24 | 47.28 | 16.90 |
| | *i-DeLVM*$_{\text{Task}}$ | **41.74** | **48.14** | **16.90** | **39.81** | **48.53** | **17.18** |
| | VQ-GAN Oracle | 10.47 | 64.92 | 24.13 | 10.47 | 64.92 | 24.13 |

Table 3: Interactive super-resolution performance in different metrics on ADE20K across different interaction cues.

### 5.1.3 Interactive Object Removal

When following an interaction cue to remove an object, we see in Table 4, that *i-DeLVM*$_{\text{Unseen}}$ is able to produce the best scores in terms of LPIPS, SSIM and PSNR, across all unseen interactions when compared to the state-of-the-art visual in-context learning models. This indicates, that *i-DeLVM*$_{\text{Unseen}}$ can consistently generalize to unseen cues that indicate the object to be removed and successively in-paint coherent background textures. Here, *i-DeLVM*$_{\text{Unseen}}$ can even reach up to the *i-DeLVM* variants which have seen the interactions in training, indicating that interactive object removal is well addressed.

|  | | Bounding Box | | | Ellipse | | | Scribbles | | | Click | | | ± Clicks | | |
|---|---|---|---|---|---|---|---|---|---|---|---|---|---|---|---|---|
|  | | LPIPS↓ | SSIM↑ | PSNR↑ | LPIPS↓ | SSIM↑ | PSNR↑ | LPIPS↓ | SSIM↑ | PSNR↑ | LPIPS↓ | SSIM↑ | PSNR↑ | LPIPS↓ | SSIM↑ | PSNR↑ |
| *Generalist* | Copy query | 31.62 | 50.73 | 15.98 | 33.05 | 50.07 | 15.98 | 25.53 | 54.27 | 17.33 | 24.28 | 54.61 | 17.47 | 27.09 | 53.93 | 17.13 |
| | DeLVM (300M) | 57.42 | 25.67 | 9.82 | 59.24 | 23.64 | 9.48 | 51.24 | 27.95 | 10.92 | 48.59 | 30.75 | 11.61 | 52.72 | 27.19 | 10.72 |
| | DeLVM (1B) | 33.59 | 49.06 | 15.50 | 35.38 | 48.29 | 15.42 | 27.89 | 52.53 | 16.79 | 25.97 | 53.48 | 17.08 | 28.18 | 52.98 | 16.85 |
| | LVM (3B) | 27.32 | 51.48 | 16.84 | 32.53 | 48.93 | 16.06 | 26.25 | 53.48 | 17.16 | 25.26 | 52.58 | 17.23 | 27.78 | 52.80 | 16.90 |
| | *i-DeLVM*$_{\text{Unseen}}$ | **24.18** | **55.42** | **19.30** | **25.83** | **55.43** | **19.30** | **20.30** | **57.33** | **20.09** | **20.33** | **57.23** | **20.13** | **21.38** | **57.05** | **20.13** |
| *Specialist* | *i-DeLVM*$_{\text{All}}$ | 22.38 | 56.47 | **20.03** | 23.36 | 56.03 | 19.76 | 19.59 | 57.63 | **20.42** | 19.46 | 57.37 | 20.21 | 20.46 | 57.20 | 20.21 |
| | *i-DeLVM*$_{\text{Int}}$ | 22.68 | **56.57** | 19.95 | 23.87 | 55.70 | 19.47 | 19.47 | **57.66** | 20.37 | 19.26 | 57.50 | 20.18 | 20.16 | **57.34** | 20.19 |
| | *i-DeLVM*$_{\text{Task}}$ | **22.02** | 56.45 | 20.02 | **22.78** | **56.18** | **19.88** | **19.27** | 57.63 | 20.41 | **19.00** | **57.53** | **20.33** | **19.87** | 57.30 | **20.40** |
| | VQ-GAN Oracle | 10.71 | 61.91 | 22.87 | 10.71 | 61.91 | 22.87 | 10.71 | 61.91 | 22.87 | 10.71 | 61.91 | 22.87 | 10.71 | 61.91 | 22.87 |

Table 4: Interactive Object Removal performance on RORD. Lower LPIPS values and higher scores in SSIM, PSNR are better.

### 5.2 Qualitative Results

To visualize these findings, we show qualitative predictions for the different interactive tasks in Figure 4. For interactive segmentation, the baselines do not adhere to the color scheme defined in the context set, and follow the segmentation setting from their training, *i.e.*, segmenting all objects, disregarding the context. *i-DeLVM*$_{\text{Unseen}}$ produces a segmentation directed to the target objects in the correct color coding, although it still makes mistakes for Ellipses and $\pm$ clicks. For the object removal task in the middle of Figure 4, DeLVM (300M) does not respect the context set and produces a segmentation, while DeLVM (1B) and LVM reconstruct the query. Curiously, LVM copies the person with the interaction to a different location, we hypothesize that this may be due to its exposure to video frame sequences in it's pre-training. Our *i-DeLVM*$_{\text{Unseen}}$ is able to conveniently erase the person, irrespective of the interaction cue given. For human pose estimation, we see a similar pattern, DeLVM (300M) fails, DeLVM (1B) produces a black prediction or detects multiple persons and LVM, akin to its pre-training, detects all poses. Merely *i-DeLVM*$_{\text{Unseen}}$ consistently generates a pose for the person with the cue, while for box, click and scribbles, the prediction is less articulate. As a multi-task, multi-interaction model *i-DeLVM*$_{All}$ generally produces good pose predictions.

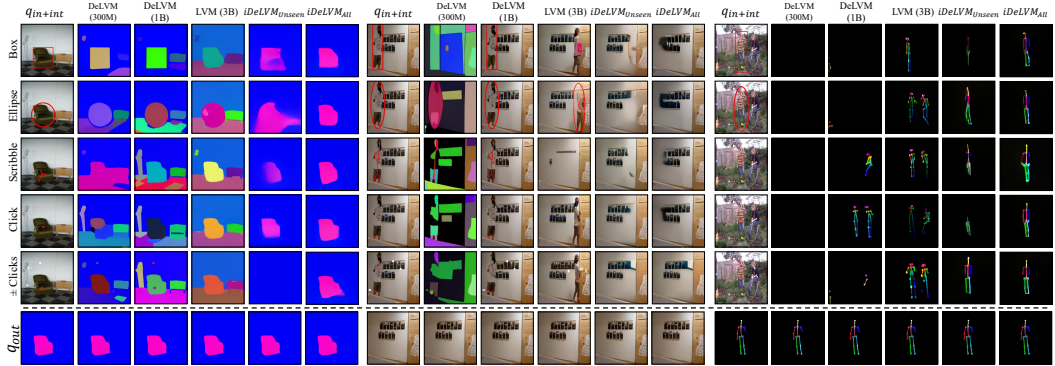

Figure 4: Qualitative results of our *i-DeLVM* and baseline models for interactive segmentation, interactive object removal and interactive human pose estimation for diverse interaction types (best viewed zooming in, context sets omitted for brevity).

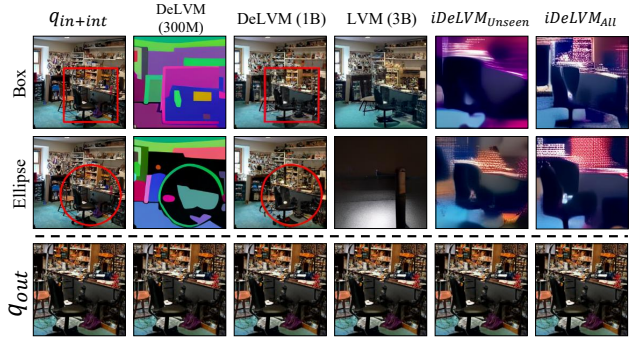

Figure 5: Qualitative results of our *i-DeLVM* and baseline models for interactive super-resolution of a chair and desk.

In Figure 5, we see that only *i-DeLVM* variants are able to zoom into the image and reproduce the coarse structures of the zoomed-in region, while baselines fail completely. Yet, the output images are very coarse with much room for improvement on such tasks, where fine details need to be recovered.

## 5.3 Ablation and Analysis

### 5.3.1 Masking ratio

We ablate the effect of conditioning the token prediction on the context $\mathcal{C}_{enc}\mathcal{Q}_{enc}$ as well as all previous tokens $\mathcal{O}^i_{enc}$ as in Equation (2), or whether masking different ratios of the previously predicted tokens can help the model to recover better from falsely predicted tokens. For interactive segmentation, using Equation (2) as is, the token accuracy is $37.86\%$, when masking $50\%$ of the previously predicted tokens, accuracy improves to $38.78\%$, while solely conditioning on $\mathcal{C}_{enc}\mathcal{Q}_{enc}$, *i.e.*, $100\%$ masking of $\mathcal{O}^i_{enc}$ leads to a token accuracy of $40.30\%$. This analysis indicates, that conditioning on $\mathcal{C}_{enc}\mathcal{Q}_{enc}$ exclusively to work best.

### 5.3.2 Random segmentation recoloring

We measure the effect of the random recoloring of segmentation maps in training. When training with fixed black and white colors for the background and target instance to segment, the IoU is $21.16\%$ and token perplexity is high at $206.90$, training with recoloring leads to a far better IoU at $56.75\%$ and $125.48$ token perplexity. As such, recoloring segmentation maps in training is an important enabling factor for context-sensitive predictions.

### 5.3.3 Interactions

To validate, that our models do not merely segment the most salient object, disregarding interactions, we prompt *i-DeLVM*$_{All}$ without interactions for segmentation. The validation IoU degrades from $52.58\%$ to merely $9.14\%$, confirming that without the cue our model can not segment the correct target object.

## 6 Conclusion

In this work, we propose the first visual in-context learning model which can address multiple interactive tasks and adapt to unseen interaction signals. We achieve this by a simple, generic and efficient encoding of interactions into the task-specific context and a training strategy based on next-token prediction. While we bridge a cruical gap in advancing static visual in-context learning models towards fluid interactivity, there remain challenges to be addressed, namely improving the performance for generative tasks such as super-resolution and enabling iterative refinement for human-in-the-loop setups. In future work, integrating our generic interaction encoding into diffusion-based models, which enable high-fidelity image generation, appears to be a promising pathway to take.

**Acknowledgment** The author acknowledges support by the state of Baden-Württemberg through bwHPC. This work was performed with the help of the Large Scale Data Facility at the Karlsruhe Institute of Technology funded by the Ministry of Science, Research and the Arts Baden-Württemberg and by the Federal Ministry of Education and Research. This work was supported by funding from the pilot program Core-Informatics of the Helmholtz Association (HGF).

## A  Datasets

We operate on the datasets PASCAL VOC (11) for interactive segmentation, the Real-world Object Removal dataset (35), the MPII dataset (1) for human poses and the ADE20K dataset (50) for super-resolution.

Several criteria influenced the selection of the dataset. Most importantly, for reproducibility, we limited the exploration exclusively to publicly available datasets. Further, for interactive visual in-context learning, we aimed to cover multiple different tasks, including semantic tasks such as segmentation and human pose estimation as well as generative tasks such as super-resolution and object removal. The selection of these tasks was done on a basis of covering well known computer vision tasks, including the well known interactive task of interactive segmentation, while also covering more exotic task setups such as interactive super-resolution, for a broad insight into in-context model performance on different tasks.

PASCAL VOC: We chose this dataset, as it is the basis for evaluation in the associated publications of DeLVM (13) and LVM (2), serving as a fitting ground for further experimentation.

Real-world Object Removal dataset: We chose this dataset, as it features besides paired images of scenes with and without objects also masks for these objects. As such, we were able to automatically derive our different notions of interactions based on these masks and integrate them into the images with objects.

MPII dataset: This dataset serves as fitting dataset for pose estimation. An alternative could have been COCO-Pose (19), yet, this dataset was already used to train both LVM and DeLVM, which would tilt the results in their favor. MPII is only used in training LVM, as such LVM has a slight edge on this task, yet, due to the excessive amount of datasets used in its training, we accept this overlap, but also emphasize that we consider a substantially different task by adding interactivity.

ADE20K dataset: Similarly, ADE20K was used in training the original DeLVM and LVM, yet it was used to train for semantic segmentation. As we use the dataset for our interactive super-resolution task, which these models have not observed on this data before, a fair comparison is ensured.

## B  Evaluation Metrics

As we evaluate three different tasks, we utilize a set of metrics to evaluate them.

Intersection over Union: For the task of interactive segmentation, we resort to the most common measure used for segmentation tasks, the Intersection over Union, which measures the overlap between the predicted segmentation and the ground truth. To compute this, we first map the prediction of the in-context models, which have the shape of RGB images, at each pixel to the closest valid color in the segmentation image. Then, the prediction is in a coherent format in which it can be compared to the segmentation ground truth.

Learned Perceptual Image Patch Similarity: For tasks where perceptual quality is a target property, *i.e.*, for the interactive object removal task and the interactive super-resolution task, we use the Learned Perceptual Image Patch Similarity. This metric indicates the distance between image patches, as such, more similarity among patches leads to lower scores. The metric is derived from network activations, which were shown to work well to quantify perceptual similarity (49).

Structural Similarity Index Measure: We use this metric to assess the quality of the prediction in the tasks interactive object removal and the interactive super-resolution. To compute the Structural Similarity Index Measure (43), we use the ground truth images without images (for object removal) and the high resolution images (for super-resolution) as distortion-free references and compare them to the predicted images of the visual in-context models.

Peak Signal-to-Noise Ratio: The tasks interactive object removal and the interactive super-resolution both contain a notion where structures from the query image need to be reconstructed. To quantify this notion Peak Signal-to-Noise Ratio is a fitting metric, which is, *e.g.*, used to evaluate image compression or reconstruction techniques.

Token Perplexity: In order to measure the accuracy of transformer models in predicting the correct tokens, perplexity is often used. In our case, a Llama model predicts token sequences from a VQ-

| System Stack | | Comment |
| --- | --- | --- |
| GPU | NVIDIA H100 / A100 | 1 GPU/run; `bfloat16` precision |
| Programming language | Python@v3.11 | Version 3.11 |
| ML library | PyTorch@v2.5.1 | |
| Training framework | PyTorch Lightning@v2.5.0 | |
| PEFT implementation | HuggingFace PEFTv@0.15.1 | `https://huggingface.co/docs/peft` |
| LLaMA/VQ-GAN implementation | InternLM | Following (13) |
| LLaMA training hyperparameters | | |
| Masking ratio | 100% | of target tokens |
| Context size | 2048 tokens | = 3 context pairs + 1 query pair |
| Batch size | 8 | Effective batch size: 16,384 tokens |
| Initial learning rate | $2.8E-4$ | |
| Learning rate scheduler | Cosine Annealing $T\_max = 50$ $eta\_min = 1e-5$ | PyTorch implementation |
| Parameter precision | 16-bit floating point | Analogous to pre-training |
| Total parameters | 299,504,640 | |
| LoRA hyperparameters | | |
| LoRA Rank r | 32 | |
| LoRA $\alpha$ | 32 | |
| Target modules | Self-Attention $[Q, V]$-projections | |
| dropout | 0.1 | |
| LoRA-trainable parameters | 2,883,584 | 0.96% of total |
| VQ-GAN hyperparameters | | |
| Resolution | 256x256 pixels | |
| Codebook Size | 8,192 | |
| Downsampling factor $f$ | 16 | |
| Resulting tokenized image size | 256 tokens | |
| Total parameters | 146,244,675 | Frozen during training |

Table 5: Model and fine-tuning hyperparameter settings used throughout the experiments.

GAN codebook. The expected predicted token sequences given a trained model can be quantified as the exponentiation of the average negative log-likelihood of the token sequence. This metric has the advantage, that it can be directly evaluated on the predicted token sequence and does not require loading the VQ-GAN model into memory for decoding the token sequence into image space.

Token accuracy: Further, the proportion of correctly predicted tokens can be quantified through the Token accuracy metric. Here, the accuracy is computed in the context of the VQ-GAN codebook, *i.e.*, whether in a token sequence the correct entries of the $N = 8,192$ tokens were predicted. This metric has the advantage, that it can be directly evaluated on the predicted token sequence and does not require loading the VQ-GAN model into memory for decoding the token sequence into image space.

## C Hyperparameter settings

In Table 5 we show the hyperparameter setting we used for fine-tuning the DeLVM Llama model (13) with LoRA (14). These hyperparameter settings were found by comparing validation segmentation results for different learning rate schedules (constant, saw-tooth, cosine annealing), learning rates and the LoRA rank parameters. The configuration that lead to the highest validation score was selected. We further document the pre-trained VQ-GAN version we used as well as the system stack and libraries in Table 5.

## D Dataset pre-processing

Next, we outline pre-processing steps for using the images of the respective datasets in our setup.

| Color | Red | Green | Blue |
|---|---|---|---|
| black | 0.000 | 0.000 | 0.000 |
| white | 1.000 | 1.000 | 1.000 |
| red | 1.000 | 0.000 | 0.000 |
| green | 0.000 | 1.000 | 0.000 |
| blue | 0.000 | 0.000 | 1.000 |
| yellow | 1.000 | 1.000 | 0.000 |
| pink | 0.984 | 0.059 | 0.750 |
| cyan | 0.000 | 1.000 | 1.000 |
| orange | 0.816 | 0.523 | 0.000 |
| brown | 0.586 | 0.293 | 0.000 |
| purple | 0.625 | 0.125 | 0.938 |
| gray | 0.500 | 0.500 | 0.500 |

Table 6: RGB values of twelve colors for recoloring augmentation of semantic maps.

### D.1 PASCAL VOC

The dataset comes with $2,913$ images that include pixel-level segmentation maps, which contain a total of $6,906$ different annotated objects to be segmented. As a result, we extract a segmentation map including each object exclusively, leading to $6,906$ single-object segmentation maps. We subsequently split this set into training, validation and test sets with a $80\%/10\%/10\%$ ratio. Crucially, we split on an image level, meaning that if an image contains more than one object, we group all the associated single-object segmentation maps into the same split in order to have isolated training, validation and test sets. The average number of objects per image is 2.38 for the training and validation split, and 2.37 for the test split, with each object occupying approximately $31\% \pm 4.2\%$ of the total image area.

We construct several single-object segmentation map variants. As basis, for each object, we extract its segmentation map and declare everything else in the image as background. We then apply random colors to the background and foreground object, which we sample from the set of twelve different colors in Table 6. This random coloring of the segmentation maps forces models trained on it to consider the context set closely, including its color scheme, potentially enhancing its ability to learn and generalize from the context examples. We describe the effects of this in an ablation experiment in the main paper.

In order to not having to load the VQ-GAN tokenization model into GPU memory when training *i-DeLVM*, we pre-extract the token sequences ahead of training. Therefore, we utilize the VQ-GAN model specified in Table 5. As this model operates on $256 \times 256$ RGB images, we resize all images and single-object segmentation maps to this size. Before tokenization of the images, we add the different interaction cues, *i.e.*, based on the segmentation we add boxes, clicks, scribbles or ellipses and afterwards tokenize the blended images.

In Figure 6 the frequency of the top 10 tokens in the tokenized dataset is displayed. Specifically, we analyze either the standard black and white coloring of segmentation maps versus the randomly re-colored segmentation maps in token space. The more heavily tilted token distribution for black and white segmentation maps could be a hint why training with re-colored segmentation maps works much better as seen in the ablation of the main paper. A reason for this might be that an imbalanced target token distribution paired with the use of the cross-entropy loss could lead to difficulties finetuning the model. For instance, Raunak *et al.* (28) showed that on a token and sequence level, low-frequency tokens are not handled as well by a transformer model as higher-frequency ones in the context of machine translation. As such, we utilize the re-colored segmentation maps for our experiments, validated by the ablation study in the main paper.

### D.2 MPII dataset

To avoid conflicts with the pre-trained model for which Guo *et al.* (13) used the COCO-Pose dataset, we use the MPII dataset by Andriluka *et al.* (1) for our experiments. It contains $24,984$ images. As for visual in-context learning, poses are encoded as RGB images, poses of persons far away from the camera are hard to encode, as all joints and their connections occupy a very small region. Due to this, we filter out the smallest $15\%$ of instances based on their bounding box leading to $17,280$ images with pose annotations. As in interactive pose estimation, the interaction cue steers the pose of which person should be predicted, we extract the poses of each individual in the images, leading to $28,561$ singular poses. We split these samples into train, validation and test sets with a split ratio of 80%/10%/10%, in case an image contains multiple individuals, we put all corresponding samples into the same split.

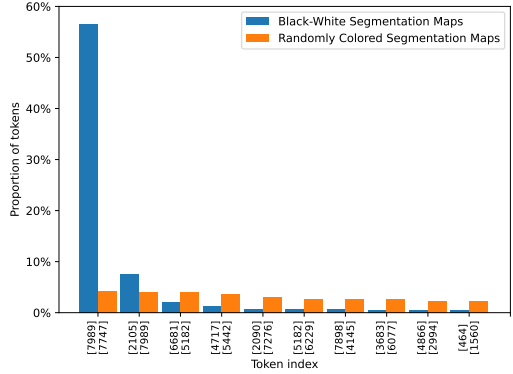

Figure 6: Distribution of top 10 tokens for white/black segmentation maps compared to colored segmentation maps.

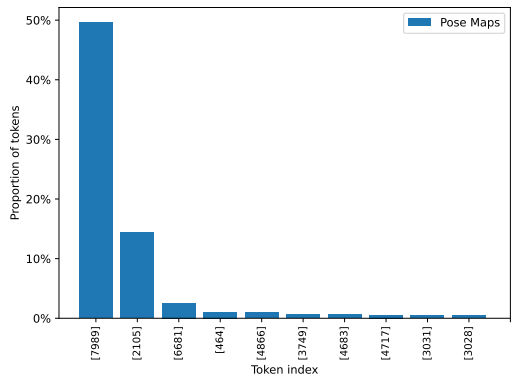

Figure 7: Distribution of top 10 tokens for pose maps on the mpii dataset.

The human poses are encoded as images using the up to $16$ keypoints, depending on potential occlusion of some parts of the person in the scene. We create the pose images by using a fixed color mapping and limb connection pattern, similar to the procedure that Guo et al. (13) used for pre-training. Specifically, we model each keypoint as a dot with a dynamically scaling radius, with a minimum size of $5$ pixels and put it onto a black image. For a pose-skeleton in the image, these keypoints are connected by lines, with a line width equal to the dot radius. The colors we utilize for the skeletons are black, red, green, blue, cyan, yellow and pink and associated to the different human joints as denoted in Table 7. As for PASCAL VOC, the input images are blended with the different interaction signals. Both images with the blended interactions and pose maps are tokenized in the same procedure as outlined for PASCAL VOC. In Figure 7 we see a long-tailed token distribution when visualizing the top 10 most frequent tokens in the pose maps, which is quite similar to the token distribution for black and white segmentation maps.

### D.3 ADE20K

On the ADE20K, we employ an adaptation of the super-resolution task where we aim to recover high-resolution images from low-resolution images. To model the degradation of an image as low resolution, we use a basic bilinear interpolation down-sampling method[1] to keep the quality loss at an equal and deterministic level.

| Keypoint | Red | Green | Blue |
|---|---|---|---|
| right ankle | 0.000 | 1.000 | 0.000 |
| right knee | 0.000 | 1.000 | 0.000 |
| right hip | 0.000 | 1.000 | 0.000 |
| left hip | 0.000 | 1.000 | 1.000 |
| left knee | 0.000 | 1.000 | 1.000 |
| left ankle | 0.000 | 1.000 | 1.000 |
| pelvis | 0.984 | 0.059 | 0.750 |
| thorax | 0.984 | 0.059 | 0.750 |
| upper neck | 0.984 | 0.059 | 0.750 |
| head top | 1.000 | 1.000 | 0.000 |
| right wrist | 1.000 | 0.000 | 0.000 |
| right elbow | 1.000 | 0.000 | 0.000 |
| right shoulder | 1.000 | 0.000 | 0.000 |
| left shoulder | 0.000 | 0.000 | 1.000 |
| left elbow | 0.000 | 0.000 | 1.000 |
| left wrist | 0.000 | 0.000 | 1.000 |

Table 7: Colors used for keypoints and skeleton images for human pose estimation.

We incorporate interactivity into the super-resolution task by enabling the user to select a portion of an image to enhance its resolution, rather than the full image. Since the model must be aware of the extent of the area to be enhanced, the interaction signals need to enclose the target area. Thus, we use bounding boxes and ellipses from our set of interactions for this task.

The pre-processing pipeline for ADE20K as super-resolution dataset that we use is to take the high-resolution images, resizing them to $256 \times 256$ pixels, which is the input size of *i-DeLVM*, adding the interaction signal which yields the input image. For the target image, we can use the corresponding cutout of the original high-resolution image based on the interaction.

The ADE20K dataset contains $27,574$ images, we filter out images smaller than $300$ pixels on the shortest side, which results in $19,019$ images for the training, validation, and test sets. We again divide these samples using an $80\%/10\%/10\%$ split. Subsequently, for each image, we select a $256 \times 256$ pixel region that encompasses between $9\%$ and $73\%$ of the image as the target. For all samples, the resulting average relative size of a cutout area is $13.06\%$.

The interactive super-resolution task has a natural image as target, making its output tokens much more diverse than the previous two tasks of segmentation and pose estimation. This can be observed in Figure 8: The most frequent tokens each only make up roughly $0.2\%$ of all the tokens, as opposed to $50-60\%$ for segmentation and pose estimation.

### D.4 Real-world Object Removal dataset

We use the dataset of Sagong *et al.* (35) for the task of object removal from scenes. It consists of real-world photographs of $3,447$ unique scenes each with a resolution of $1920 \times 1080$ pixels. Within each scene, one or more objects are captured in different positions across multiple frames along with an additional image that excludes these objects. Importantly, in each frame, object masks are available for all objects that are removed in the target image, which enables us to supply them with our interactions. The dataset provides roughly equal amounts of objects with sizes between $0\%-10\%$, $10\%-20\%$, $20\%-30\%$ and greater than $30\%$ of the whole image.

Each scene consists of 150 frames on average, out of which we randomly sample 6 to adjust the dataset size to the size of the other tasks. We subsequently split these images into training, validation and test data, as with the other tasks with a split ratio of $80\%/10\%/10\%$. In the samples, multiple objects can be present in a given frame in different positions, thus requiring multiple interaction signals for each differently positioned object.

---

[1]https://docs.pytorch.org/vision/stable/transforms.html

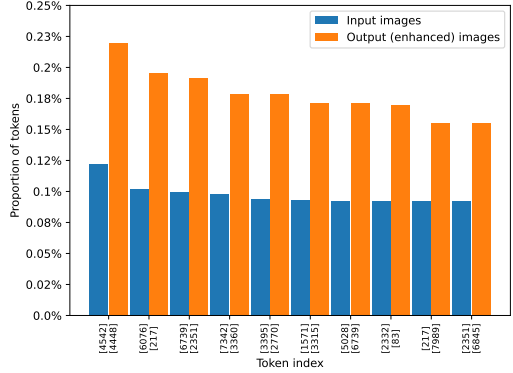

Figure 8: Distribution of top 10 tokens for interactive super-resolution on the ADE20K dataset.

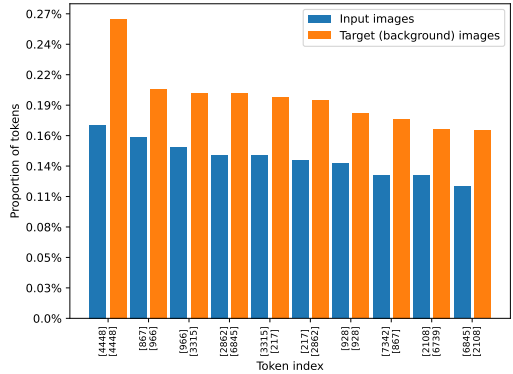

Figure 9: Distribution of top 10 tokens for interactive object removal on the RORD images.

As with the interactive Super-resolution task, the targets of the object removal task are photorealistic images, making the token space more evenly distributed than for segmentation- and pose maps. Figure 9 shows this by presenting the 10 most frequent tokens from the tokenized input images and corresponding targets which show the input image without the objects. The pre-trained VQ-GAN model that we use in our experiments achieves a $63.23\%$ SSIM, $22.37 dB$ PSNR, when reconstructing images from the dataset. In Figure 10, we show example images for each task and interaction cue.

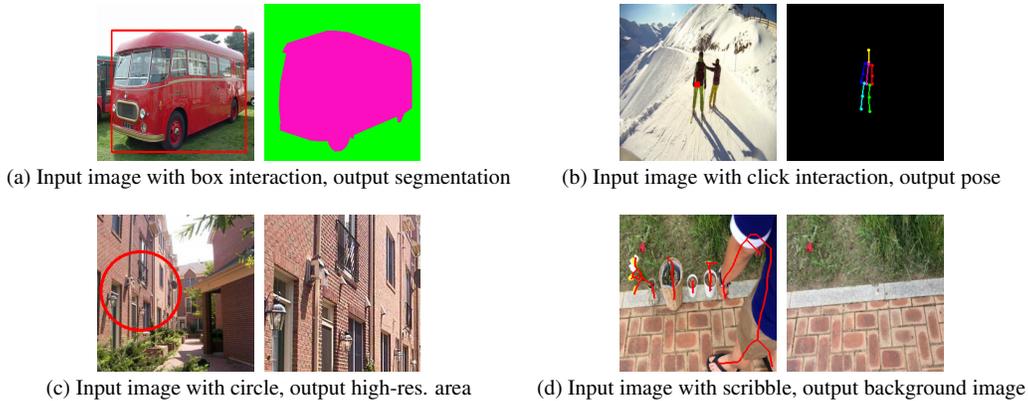

(a) Input image with box interaction, output segmentation  (b) Input image with click interaction, output pose

(c) Input image with circle, output high-res. area  (d) Input image with scribble, output background image

Figure 10: Example pairs from the different datasets for each task with different interaction cues blended into the input image.

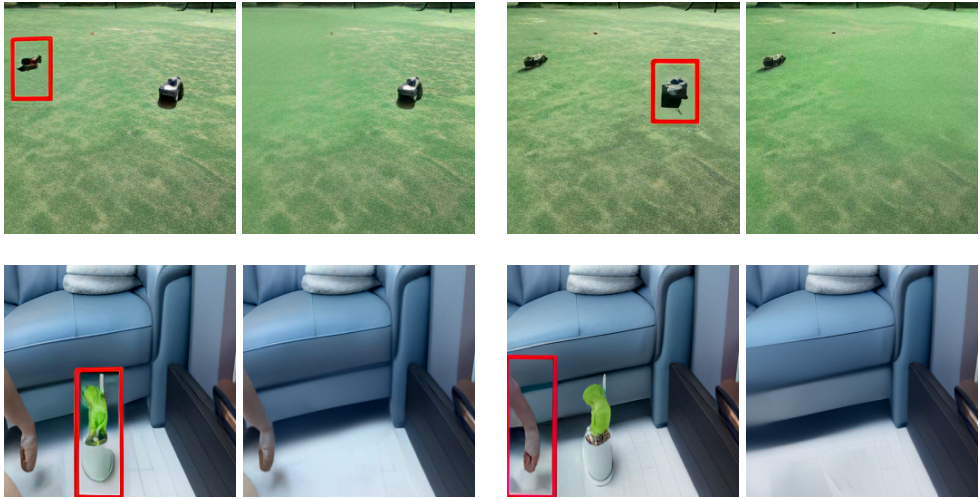

Figure 11: Object removal success and failure cases. For each pair, the images to the left are the queries and to the right are the model's predictions. In all predictions, the *i-DeLVM$_{All}$* model variant was used.

## E    Interactions for Object Removal

The fact that all prominent objects are removed at once in training of *i-DeLVM* could lead to trained models ignoring the interaction signals and removing all salient objects regardless of the interactions. To ensure that the interactions are considered by our *i-DeLVM* models, we constructed a testing scenario where only one of the objects to be removed is selected with an interaction cue, while the others are retained. In Figure 11, we see some qualitative results, which indicate, that singular objects can indeed be removed (row one). In row two, we see another success case to the left, where the object is correctly removed, while the right pair of images shows a failure case, where all objects are removed, including a neighboring object that was not interacted with.

## F    Masking ratio graph over training epochs

In the ablation studies of the main paper, we explore the effect of adapting the auto-regressive next-token prediction loss formula. Specifically, we successively mask out more and more conditional output tokens $\mathcal{O}^i_{enc}$ from Equation (2) of the main paper. In Figure 12, we further provide a graph over the training steps, to see the effect of training with either no masking, $50\%$ masking or $100\%$ masking during trianing. We measure the token accuracy at different steps on the validation set, where we see that a higher masking ratio leads to better results. As we outline in the main paper, therefore, all our models in the main results are trained with a masking ratio of $100\%$ which corresponds to conditioning the model exclusively on the context tokens $\mathcal{C}_{enc}\mathcal{Q}_{enc}$ when predicting the next token, and dropping the conditioning on previously predicted output tokens.

## G    Autoregressive token-by-token vs. single-forward pass prediction

In Table 8, we compare whether results change drastically if we deviate from predicting the output token-by-token, *i.e.*, with one forward pass per token, towards predicting all tokens at once in a single forward pass. For the token-by-token scenario, we use greedy decoding as the generation strategy: For each forward pass, the token with the highest logit score is predicted and appended to the sequence, which serves as input in the following next-token prediction, repeating until $256$ tokens are predicted, which equates to the tokens of one full image.

Predicting each token given the original input and previously predicted tokens almost consistently outperforms the single-forward-pass strategy, but not by particularly large margins. For the object removal task both token-by-token and single-forward pass perform roughly the same. This means that with slightly diminished returns, inference is possible with a fraction of the compute and time ($\frac{1}{256}$) by generating the whole task-specific output image in one forward pass. However, in the experiments of the main paper, we use token-by-token inference for better results.

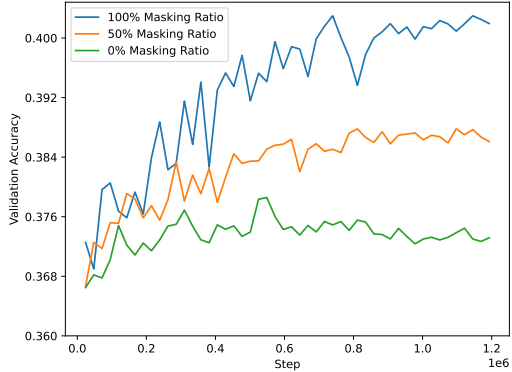

Figure 12: Throughout the training steps, a higher masking ratio leads to higher token accuracy on the interactive segmentation token sequences.

| Task | Metric | Token-by-token | Single-forward |
|---|---|---|---|
| Segmentation | IoU (%) | **47.09** | 46.79 |
| Pose | IoU (%) | **9.56** | 9.35 |
| Enhance | SSIM (%) | **47.13** | 46.48 |
| Object removal | SSIM (%) | 56.938 | **56.939** |

Table 8: Scores for the token-by-token and single-forward-pass inference strategies as an average over all interaction types.

## H  Construction of interactions

For box interactions, we draw red boxes around the provided pixel-wise masks, similarly we draw ellipses which fully encompass the masks. An exception is the interactive super-resolution task, where we freely choose the size and location of these two interaction cues. For scribbles, we use the skeletonize function from the scikit library[2] on the pixel-wise masks and thicken them through morphology operations using the same library. Clicks are computed as the center of gravity of the pixel-wise mask which still lies within the mask. Positive and negative clicks are computed at random locations around the target instance.

## I  Calculating reconstruction quality of interactions

In the main paper, in the section titled *'Manifestation of Interactions in Token Space'*, we evaluate the retainment of interactions when converting them into token space and back into image space. To this end, an image containing an interaction is auto-encoded and later post-processed in the following way: First, the image is cropped to the interaction region, with a tolerance of 2 pixels to each side. For bounding boxes, the content within the box is removed by setting the pixels to $0$. After this, the cropped region is filtered for red values (if the specified region in the image already contained red color, it is discarded) by first transforming the image into HSV color space and removing each pixel if its hue value is farther away than $10°$ from $360°$ or its saturation or value are less than $\sim 39\%$. These thresholds were defined based on empirical evaluation of reconstructed interactions. Finally, the filtered reconstruction is compared with the original interaction, where we calculate the overlap with the Intersection over Union metric.

---

[2]https://scikit-image.org/docs/stable/api/skimage.morphology.html

## J Interaction effects on token space

The effect of superimposed interactions on images can be observed by looking at how many tokens change when adding the interactions and comparing the original tokenized image with the tokenized image with superimposed interactions. We do this for images in the PASCAL VOC dataset, encode the images, and check how many tokens in the vicinity of the interaction are affected by comparing the token maps with their non-interacted counterparts. Figure 9 in the main manuscript shows the results of this experiment for bounding boxes with different sizes. We extract samples where the sides are of similar length and group them in buckets of four. This means, for instance, bounding boxes with side lengths of $4 - 7$ pixels are grouped. We later extract only an prototypical subset of these groupings for the final visualization, so the bounding box sizes in Table 9 are the buckets (4, 7), (32, 35), (60, 63), (88, 91), (116, 119), (144, 147), (172, 175), (200, 203), (228, 231), respectively. The images attribute brighter pixels at locations with frequent token changes, black values where tokens do not change. As expected, the highest influence of the interaction on the token space occurs in it's direct neighborhood. Small boxes directly influence tokens in a nucleus shape around them, while larger boxes tend to change the content they encompass less. Although it has to be said, that a token change is not necessarily bad, as it does further indicate the presence of the interaction cue in this latent token space and it may still allow for faithful image content reconstruction.

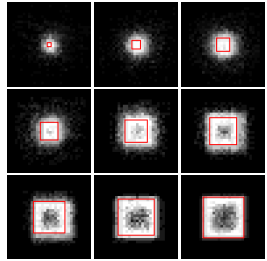

Table 9: Changed tokens when bounding box is added with different sizes.

## K Qualitative figures in higher resolution

In Figure 13, 14, 15 and 16, which can be found on the subsequent pages, we display the qualitative results from the main paper again in higher resolution for better inspection. Additionally, in these figures, we show example context sets with the bounding box interaction. For the qualitative results with other interactions, context sets with their respective interaction type are used, for brevity and visual clarity, we omit displaying all of them.

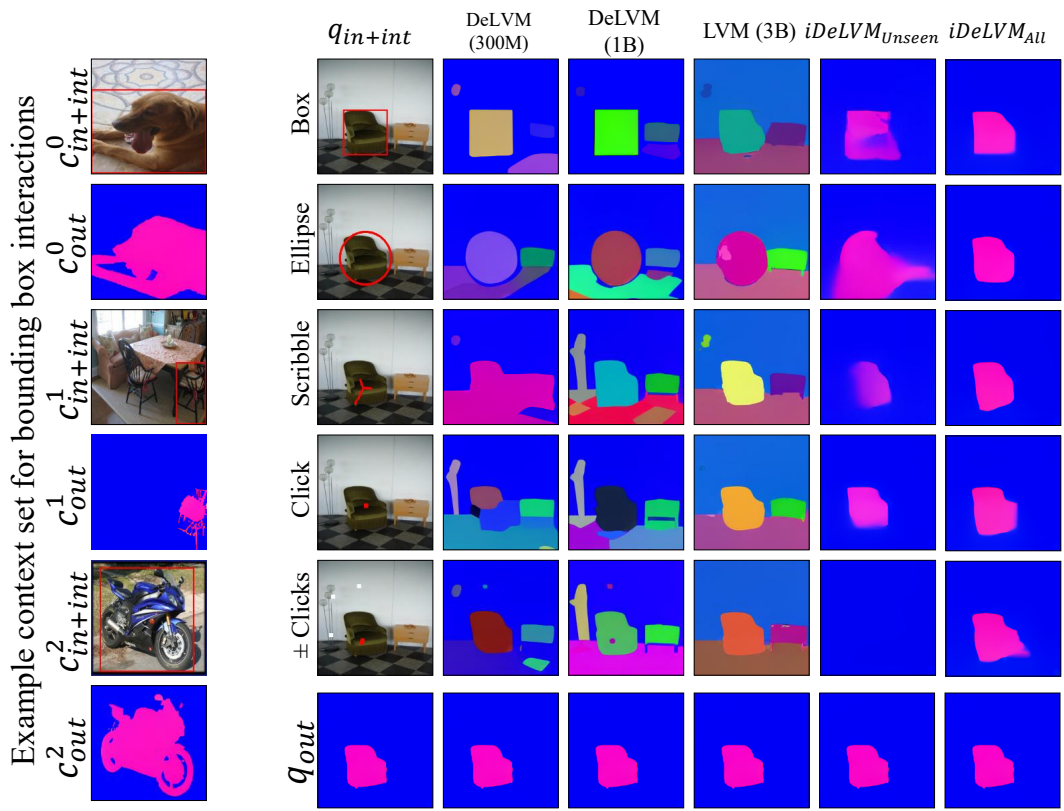

Figure 13: Qualitative interactive segmentation results from the main paper in higher resolution for better inspection.

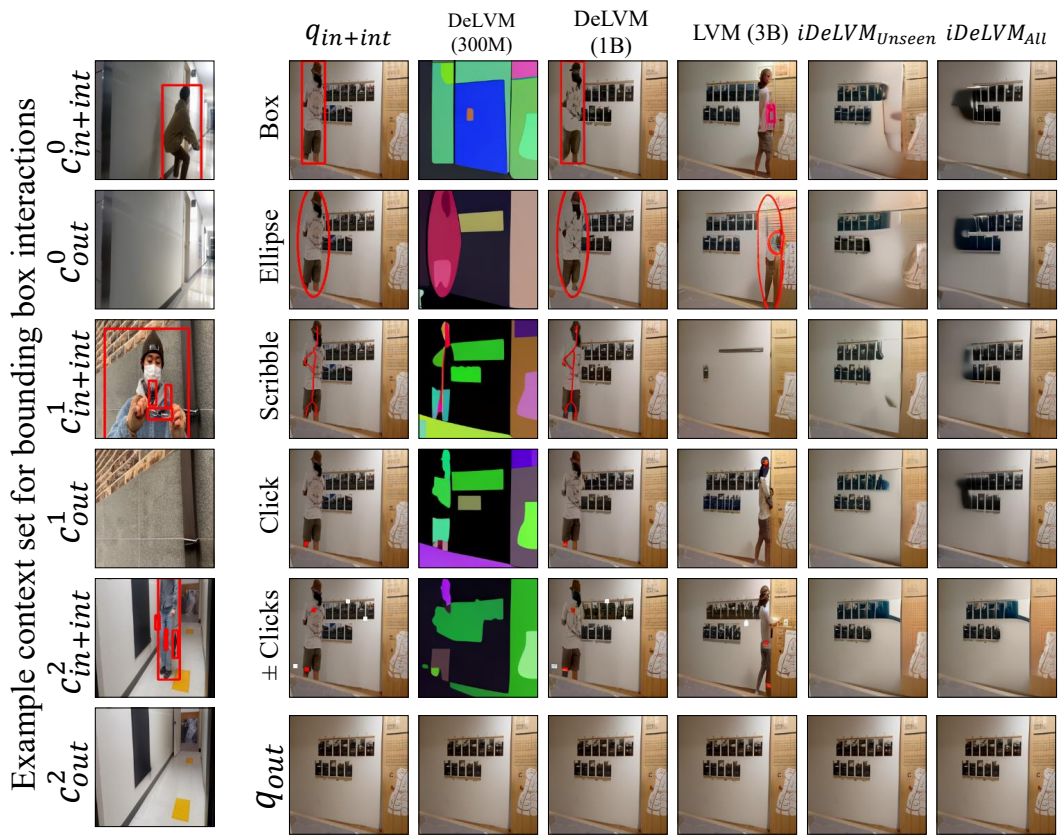

Figure 14: Qualitative interactive object removal results from the main paper in higher resolution for better inspection.

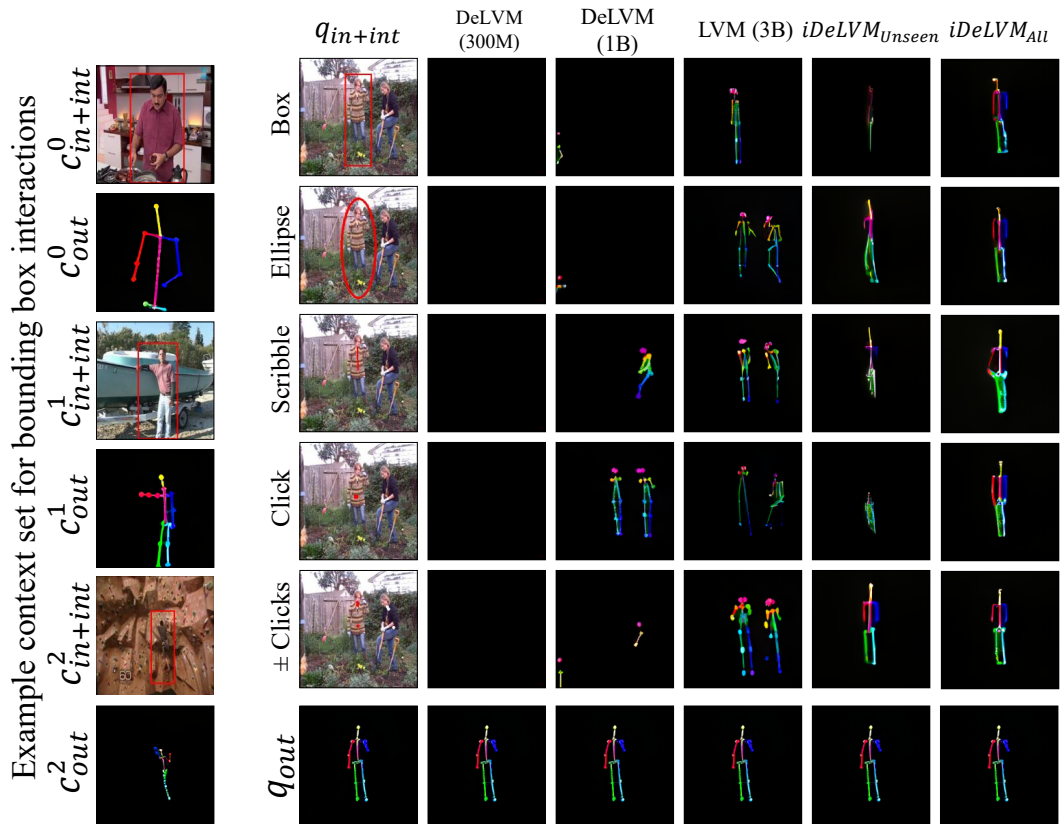

Figure 15: Qualitative interactive pose estimation results from the main paper in higher resolution for better inspection.

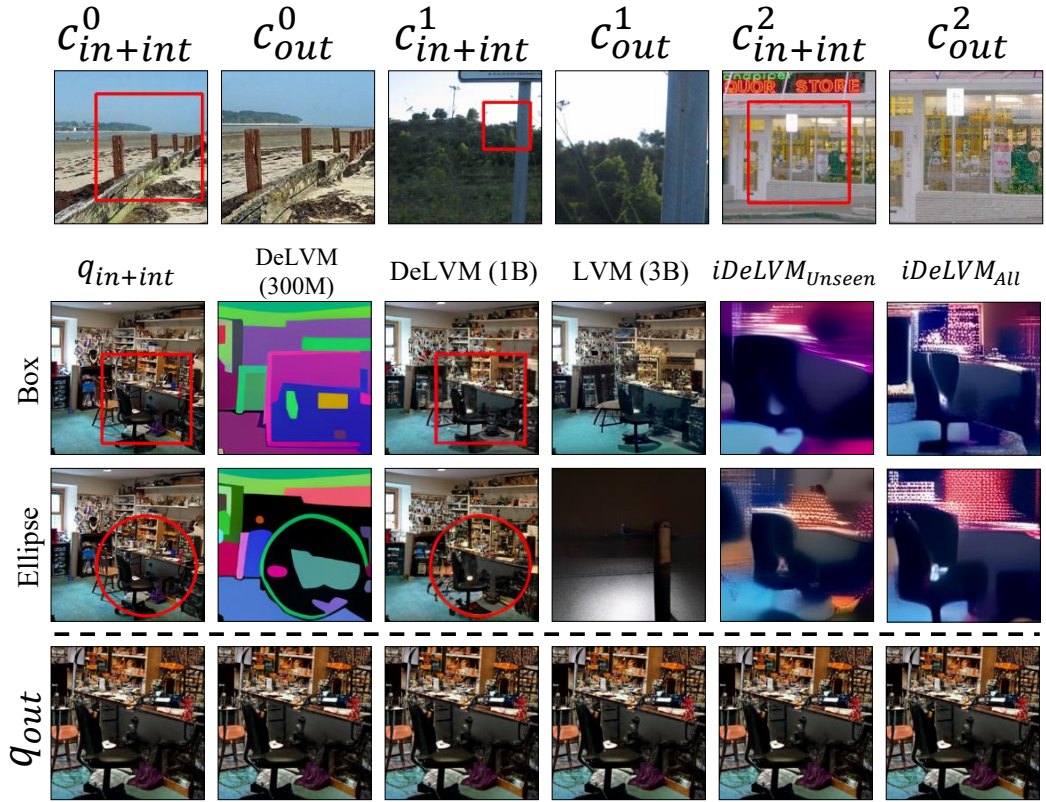

Figure 16: Qualitative interactive super-resolution results from the main paper in higher resolution for better inspection.